\documentclass[letterpaper, 10 pt, journal, twoside]{IEEEtran}
\IEEEoverridecommandlockouts

\usepackage{graphics}
\usepackage{epsfig}
\usepackage{mathptmx}
\usepackage{times}
\usepackage{amsmath}
\usepackage{amssymb}
\usepackage{cite}
\usepackage{booktabs}
\usepackage{multirow}
\usepackage{pifont}
\usepackage{color}
\usepackage{bookmark}
\usepackage{hyperref}
\hypersetup{colorlinks=true}
\newcommand{\cmark}{\color{green}\ding{52}}
\newcommand{\ymark}{\color{yellow}\ding{52}}
\newcommand{\xmark}{\color{red}\ding{54}}

\title{
VECtor: A Versatile Event-Centric Benchmark\\for Multi-Sensor SLAM
}

\author{Ling Gao$^{*}$, Yuxuan Liang$^{*}$, Jiaqi Yang$^{*}$, Shaoxun Wu, Chenyu Wang, Jiaben Chen, and Laurent Kneip
\thanks{Manuscript received: February, 24, 2022; Revised May, 19, 2022; Accepted June 15, 2022. This letter was recommended for publication by Editor Sven Behnke upon evaluation of the Associate Editor and Reviewers' comments. This work was supported by the Natural Science Foundation of Shanghai (grant number 22ZR1441300), as well as the generous support provided by our industry partner Stereye Intelligent Technology.}
\thanks{* denotes these three authors contributed equally.
        The authors are with the Mobile Perception Lab of the School of Information Science and Technology, ShanghaiTech University (\protect\url{https://mpl.sist.shanghaitech.edu.cn}).
        L.~Gao is also affiliated with Shanghai Institute of Microsystem and Information Technology, Chinese Academy of Sciences, and University of Chinese Academy of Sciences.
        Y.~Liang is now with Northwestern University, United States.
        L.~Kneip is also affiliated with the Shanghai Engineering Research Center of Intelligent Vision and Imaging.}
\thanks{Digital Object Identifier (DOI): see top of this page.}
}

\begin{document}

\maketitle


\begin{abstract}
Event cameras have recently gained in popularity as they hold strong potential to complement regular cameras in situations of high dynamics or challenging illumination. An important problem that may benefit from the addition of an event camera is given by Simultaneous Localization And Mapping (SLAM). However, in order to ensure progress on event-inclusive multi-sensor SLAM, novel benchmark sequences are needed. Our contribution is the first complete set of benchmark datasets captured with a multi-sensor setup containing an event-based stereo camera, a regular stereo camera, multiple depth sensors, and an inertial measurement unit. The setup is fully hardware-synchronized and underwent accurate extrinsic calibration. All sequences come with ground truth data captured by highly accurate external reference devices such as a motion capture system. Individual sequences include both small and large-scale environments, and cover the specific challenges targeted by dynamic vision sensors.
\end{abstract}



\section*{MULTIMEDIA MATERIAL}

The dataset, along with the documentation and the toolbox, can be found at \protect\url{https://star-datasets.github.io/vector/}


\section{INTRODUCTION}

\IEEEPARstart{S}{imultaneous} Localization And Mapping (SLAM) is regarded as an essential problem to be solved by intelligent mobile agents such as autonomous robots, XR devices, and smart vehicles. LiDARs or depth cameras provide direct depth readings, and are therefore often considered very helpful in reducing the complexity and increasing the accuracy and density of a SLAM solution. However, a compact form factor, low energy consumption, and the ability to sense appearance information have since ever made regular cameras an indispensable addition to any SLAM sensor suite. The present letter targets the addition of yet another exteroceptive visual sensor: a Dynamic Vision Sensor (DVS), also called an \emph{event camera}.

\begin{figure}[t]
  \centering
  \includegraphics[width=0.485\textwidth]{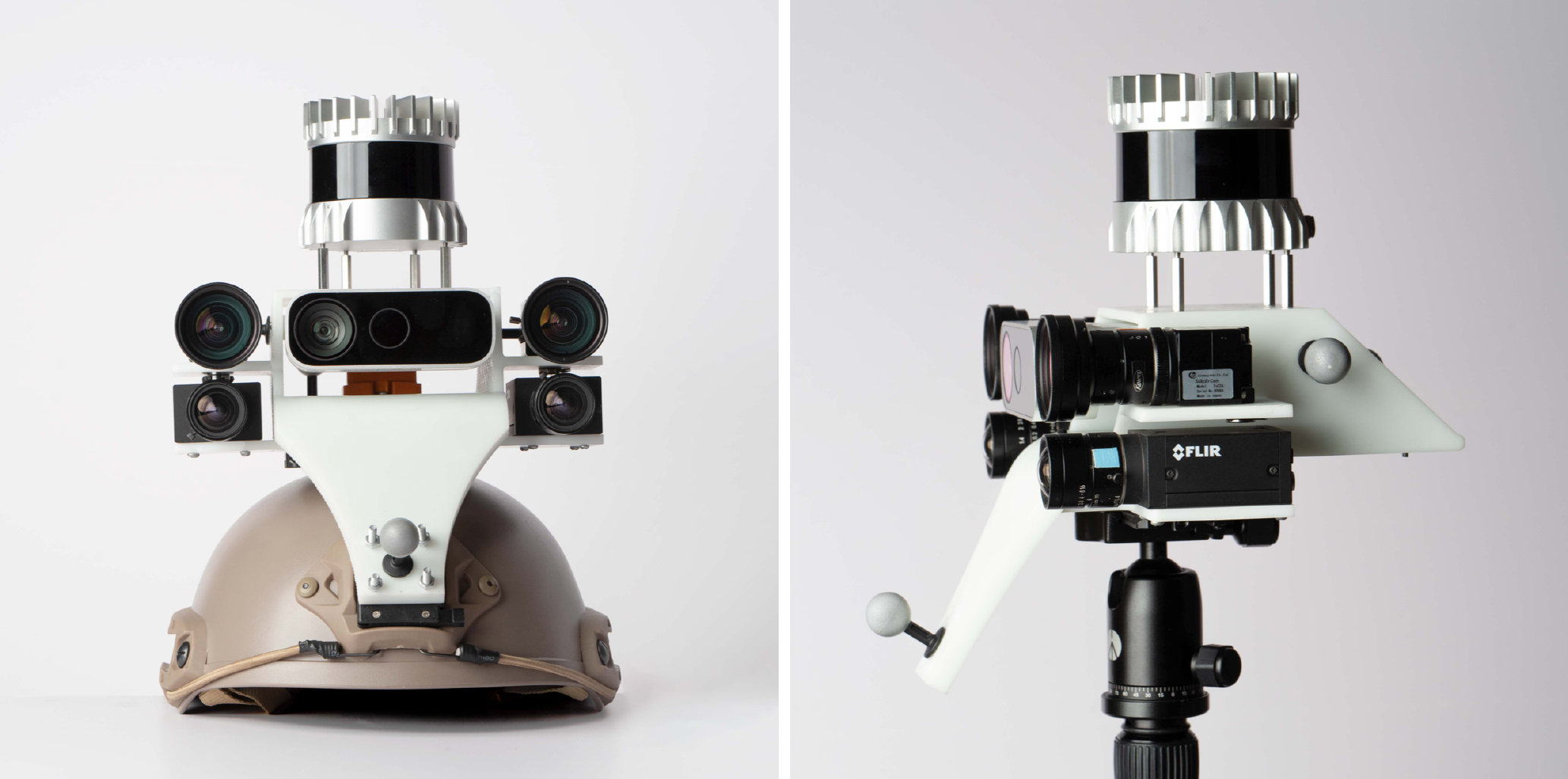}
  \caption{We present a comprehensive set of data sequences which is recorded by a rich sensor setup placed on a versatile 3D-printed holder. It is centered around an event-based stereo camera, and further contains a regular stereo camera, an \mbox{RGB-D} sensor in the center, a LiDAR mounted on the top, and an IMU at the rear. The holder can be mounted on various platforms, such as a helmet as shown on the left (front view), or on a tripod as shown on the right (side view).}
  \label{sensor_fig}
\end{figure}

\begin{figure*}[t]
 \centering
 \includegraphics[width=\textwidth]{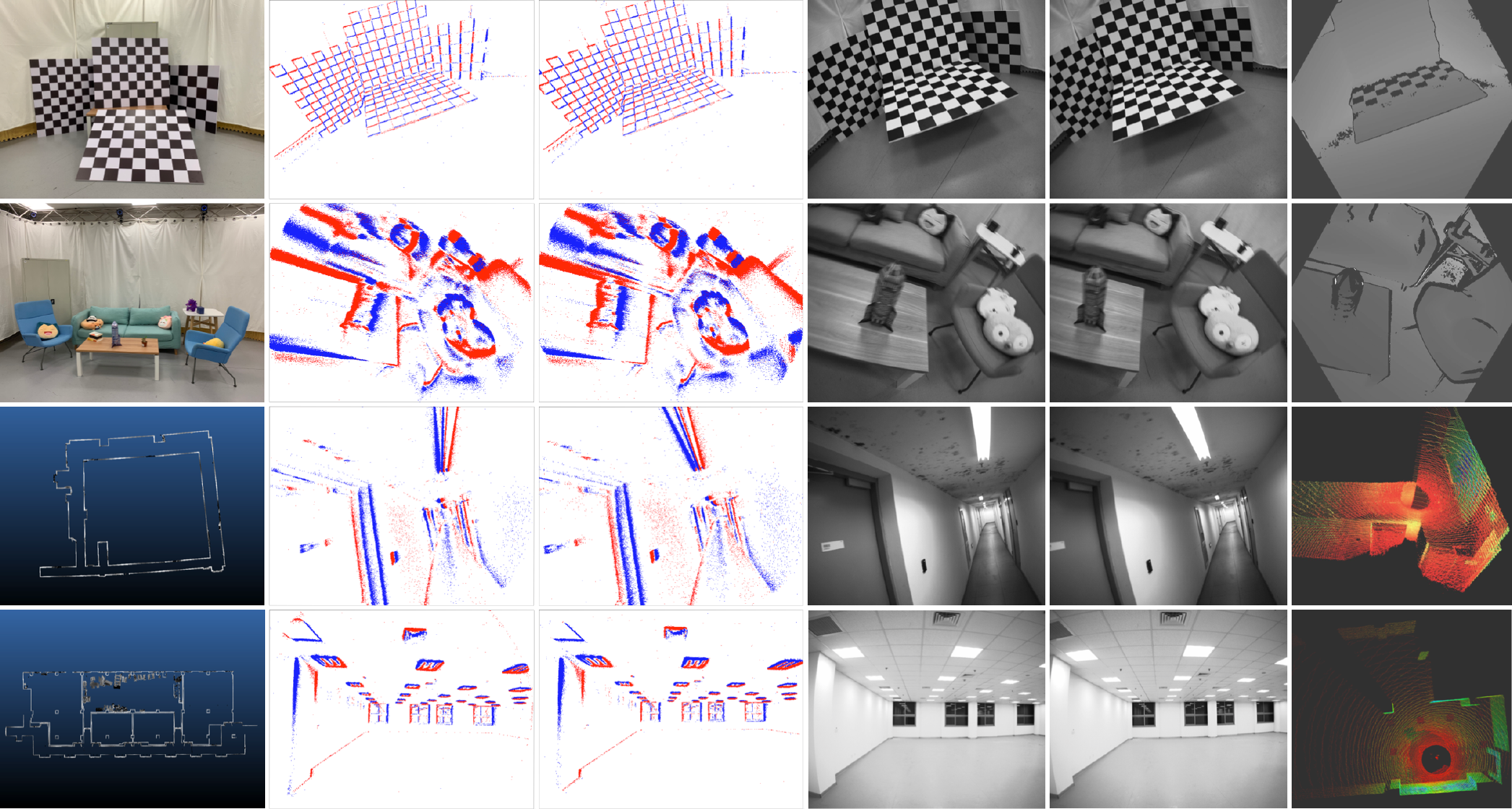}
 \caption{An overview on four selected data sequences. From top to bottom: \emph{mountain-normal}, \emph{sofa-fast}, \emph{corridors-walk}, \emph{units-dolly}. From left to right: a photo of the scene shot by mobile phone or a laser-based reconstruction of the whole environment, accumulated events for left camera (33ms), accumulated events for right camera (33ms), left visual frame, right visual frame, depth frame or LiDAR frame.}
 \label{data_seq_fig}
\end{figure*}

Event cameras have become popular since a little over a decade now. On an event camera, each pixel measures changes of the logarithmic brightness. It fires a time-stamped event as soon as the said measure changes by a certain threshold amount with respect to a reference value, and the latter is reset each time an event is fired. As a result, the measurements returned by an event camera consist of an asynchronous stream of time-stamped and localized events indicating discretized changes of the image intensity. Owing to high temporal resolution and sensitivity, the stream of events tends to be rather dense in time, while their elevated likelihood of being triggered along moving intensity edges makes their distribution in the image plane semi-dense. For further details on the internal operation principle of event cameras, the reader is kindly referred to prior art~\cite{gallego2020event, berner2013240}.

The addition of dynamic vision to a SLAM sensor suite is considered interesting from a number of perspectives. First, the high temporal resolution and absence of exposure intervals mean that event cameras have the ability to sense highly dynamic intensity changes as for example caused by the aggressive motion of a racing drone~\cite{delmerico2019we}. Second, again owing to the absence of exposure times, event cameras have High Dynamic Range (HDR) and are able to sense events in low or challenging illumination conditions, as for example caused by the simultaneous observation of both bright and dark segments. Third, the fact that the measurements are event-based enables event-driven processing, a paradigm that promises efficient perception with actively triggered calculations whenever an actual change in the observed brightness patterns occurs. Lastly, event cameras can be integrated with regular cameras on one and the same chip, and thus hold the potential to preserve the beneficial form and energy factors of passive visual sensing. The potential of adding event cameras to the set of exteroceptive sensors used in SLAM has already been demonstrated in recent works~\cite{vidal2018ultimate}.

Research and development of SLAM solutions require benchmark sequences that provide a set of standardized tests with a clean, well-calibrated hardware setup, and---most importantly---exact ground truth information. Popular examples are given by the TUM-RGBD~\cite{sturm2012benchmark}, KITTI~\cite{geiger2013vision}, and EuRoC~\cite{burri2016euroc} datasets. While simulated datasets~\cite{mueggler2017event, rebecq2018esim, hu2021v2e} certainly represent an interesting source of data, simulators are always based on models that approximate the real principles of image formation and often ignore sources of noise or certain aspects of physics and real-world camera operation. The development of multi-modal SLAM solutions that process depth, images, and events therefore asks for novel benchmark sequences captured by a real sensor setup that includes all modalities. Owing to the event cameras' ability to sense in difficult illumination and highly dynamic scenarios, the benchmark should necessarily include such challenges besides the regular small and large-scale application scenarios.

We present novel SLAM benchmark datasets which are the first to simultaneously satisfy the following requirements:

\begin{itemize}
  \item Captured by a full hardware-synchronized sensor suite that includes an event stereo camera, a regular stereo camera, an \mbox{RGB-D} sensor, a LiDAR, and an IMU;
  \item Covering the full spectrum of motion dynamics, environment complexities (e.g. basement with ambiguous geometries, unfurnished units lacking texture), and illumination conditions (e.g. HDR scenes, low or dynamically-changing illumination situations);
  \item Complete six degrees of freedom ground truth signals for both small and large-scale scenarios, and highly-accurate intrinsic and extrinsic calibration.
\end{itemize}

Note that all datasets and tools for evaluation and calibration are made publicly available, and all data sequences have been applied to a series of popular SLAM algorithms for validation purposes.


\section{RELATED WORK}

\begin{table*}[t]
  \caption{Comparison of different event-centric datasets}
  \label{ref_table}
  \begin{center}
  \begin{tabular}{cccccccccccc}
    \toprule
    \multirow{2}{*}{Dataset}            & Event           & Event  & Regular & \mbox{RGB-D}  & \multirow{2}{*}{LiDAR} & \multirow{2}{*}{IMU} & \multirow{2}{*}{Motion} 
                                        & \multirow{2}{*}{Ground Truth Poses} & \multirow{2}{*}{Sync.}                                                    \\ 
                                        & Resolution      & Stereo & Stereo & Sensor & & & & &                                                            \\
    \midrule
    D-eDVS\cite{weikersdorfer2014event} & $128\times128$  & \xmark & \xmark & \cmark & \xmark & \xmark & Hand-held     & MoCap                   & \xmark \\ 
    evbench\cite{barranco2016dataset}   & $240\times180$  & \xmark & \xmark & \cmark & \xmark & 6      & Mobile Robot  & Odometer                & \xmark \\ 
    MVSEC\cite{zihao2018multi}          & $346\times260$  & \cmark & \cmark & \xmark & 16     & 6/9    & Diverse       & MoCap + Cartographer    & \ymark \\ 
    UZH-FPV\cite{delmerico2019we}       & $346\times260$  & \xmark & \cmark & \xmark & \xmark & 6      & Drone         & Total Station w/ spline & \ymark \\ 
    ViViD\cite{lee2019vivid}            & $240\times180$  & \xmark & \xmark & \cmark & 16     & 9      & Hand-held     & MoCap + LOAM            & \ymark \\ 
    ViViD++\cite{lee2022vivid}          & $640\times480$  & \xmark & \xmark & \xmark & 64     & 9      & Driving       & LOAM w/ GPS             & \ymark \\ 
    DSEC\cite{gehrig2021dsec}           & $640\times480$  & \cmark & \cmark & \xmark & 16     & \xmark & Driving       & RTK GPS                 & \cmark \\ 
    AGRI-EBV\cite{zujevs2021event}      & $240\times180$  & \xmark & \cmark & \cmark & 16     & 6      & Mobile Robot  & LiDAR SLAM              & \ymark \\ 
    TUM-VIE\cite{klenk2021tum}          & $1280\times720$ & \cmark & \cmark & \xmark & \xmark & 6      & Diverse       & MoCap (partial)         & \ymark \\ 
    Ours                                & $640\times480$  & \cmark & \cmark & \cmark & 128    & 9      & Diverse       & MoCap + ICP             & \cmark \\ 
    \bottomrule
  \end{tabular}
  \end{center}
\end{table*}

A number of benchmarks containing events in conjunction with other sensing modalities have been released in recent years. Table~\ref{ref_table} lists all datasets with key features such as sensor setups, a spectrum of motion dynamics, the nature of the ground truth signals, and the level of sensor synchronization. The first cross-modal dataset is proposed by Weikersdorfer et~al.~\cite{weikersdorfer2014event} in 2014, providing both event ($128\times128$, events only) and \mbox{RGB-D} streams with relatively low resolution. Data sequences feature an office environment set up in a Motion Capture (MoCap) arena. In 2016, Barranco~et~al.~\cite{barranco2016dataset} presented a series of indoor data sequences captured by an Inilabs DAVIS240B sensor ($240\times180$, event and APS frames, six-axis built-in IMU) along with a Microsoft Kinect \mbox{RGB-D} sensor. The whole setup is mounted on top of a pan-tilt unit and further attached to a mobile robot, hence the motion is constrained to five Degrees of Freedom (DoF). Approximate ground truth is generated by a drift-affected integration of odometry readings.

MVSEC~\cite{zihao2018multi} is considered as the first modern cross-modal dataset given its rich sensor setup of a pair of DAVIS m346B sensors ($346\times260$, events and APS frames, six-axis built-in IMU, about 10cm baseline), a VI-Sensor that includes a stereo camera (about 10cm baseline) and an in-built nine-axis IMU, and a 16-channel LiDAR. The provided sequences can be categorized by motion type as they are recorded by a hexacopter, a hand-held device, a driving car, and on a motorcycle. Ground truth as well as depth maps in the event frames are generated by running Cartographer~\cite{hess2016real} and LOAM~\cite{zhang2014loam} respectively. The first two categories come with ground truth readings from a MoCap system. However, malfunctions occur during the indoor flying sequences where the VI-Sensor data is not available. Furthermore, sensors are only partially synchronized, and temporal post-alignment is conducted.
The UZH-FPV~\cite{delmerico2019we} datasets specialize in drone racing and contain a series of aggressive flight trajectories. A single miniDAVIS346 ($346\times260$, events and APS frames, six-axis built-in IMU) with a wide-angle lens and a stereo camera with fisheye lenses are mounted on the drone. Three DoF ground truth is collected with a total station by measuring the position of an onboard reflective prism. However, this setup constrains the drone to constantly remain in the line-of-sight of the total station and fly in a large, open area. Additionally, partial tracking failures during high-acceleration maneuvers are reported, and sensors are only partially synchronized. 
ViViD~\cite{lee2019vivid} and ViViD++~\cite{lee2022vivid} are dataset projects with handheld and driving sensor setups. The handheld sequences are recorded by a single DAVIS240C ($346\times260$, events and color APS frames, six-axis built-in IMU), an RGB camera, a thermal camera, an \mbox{RGB-D} sensor, and a 16-channel LiDAR. The driving sequences first remove the \mbox{RGB-D} sensor due to its unreliable depth readings on the outside, then exchange the event sensor and the 16-channel LiDAR for a DVXplorer ($640\times480$, events only, six-axis built-in IMU) and a 64-channel LiDAR. The hand-held sequences are captured in a MoCap arena, while the outdoor sequences contain ground truth by fusing LOAM and GPS.

Further datasets have been introduced since 2021. Similar to the MVSEC sequences, DSEC~\cite{gehrig2021dsec} uses two stereo cameras but omits the LiDAR and the IMU. 
Zujevs et~al. propose agricultural robotics-oriented datasets~\cite{zujevs2021event} that are recorded by a mobile robot in different types of agricultural environments during autumn season. Sensors include a single DVS240 ($240\times180$, events only, six-axis built-in IMU), a stereo camera, an \mbox{RGB-D} sensor, and a 16-channel LiDAR. All sensors are hardware-synchronized, except for the \mbox{RGB-D} sensor. Due to technical reasons, GPS is not included in the data, and ground truth is approximated by three different LiDAR SLAM algorithms.
Finally, the TUM-VIE~\cite{klenk2021tum} dataset is captured by a pair of Prophesee Gen4 CD event cameras ($1280\times720$, events only), a stereo camera, and a six-axis IMU. The sequences are recorded in differently scaled environments and under different motion conditions (e.g. walking, running, skating, and biking). Ground truth poses are provided by a MoCap system.

An increasing number of recent works rely on public datasets to evaluate algorithm performance. Our contribution is the first to simultaneously provide a complete sensor setup including two stereo cameras, bi-modal depth data, full synchronization, accurate extrinsic calibration, and reliable, drift-free ground truth signals. We therefore believe that our work provides strong value to the sensor fusion research community.


\section{HARDWARE SETUP}

Our sensor setup consists of a multi-camera setup equipped with a LiDAR on top and a nine-axis IMU at the rear. All sensors are rigidly mounted on a 3D-printed holder, as depicted in Fig.~\ref{sensor_fig}. This section introduces the individual sensors and all details about their synchronization and calibration.

\subsection{Sensor Setup}

\begin{table}[t]
  \caption{Hardware Specifications}
  \label{hw_spec_table}
  \begin{center}
  \begin{tabular}{ccc}
    \toprule
    Sensor                      & Rate  & Specifications \\
    \midrule
    2$\times$ Prophesee Gen3 CD &       & $640\times480$ pixels                        \\
    with Kowa LM5JCM            & N/A   & f/2.8-16, FoV: 67$^{\circ}$H / 82$^{\circ}$V \\
    baseline: 17cm              &       & down to $0.08$lux, $>120$dB                  \\
    \midrule
    2$\times$ FLIR Grasshopper3 &       & $1224\times1024$ pixels (binned)             \\
    with Kowa LM6JC             & 30Hz  & f/1.4-16, FoV: 61$^{\circ}$H / 82$^{\circ}$V \\
    baseline: 17cm              &       & monochrome with global shutter               \\
    \midrule
                                &       & $640\times576$ pixels, up to 3.86m           \\ 
    Azure Kinect                & 30Hz  & Narrow FoI: 75$^{\circ}$H / 65$^{\circ}$V    \\
                                &       & only depth camera is in use                  \\
    \midrule
                                &       & $128\times2048$ points, up to 50m            \\ 
    Ouster OS0-128              & 10Hz  & FoV: 360$^{\circ}$H / 90$^{\circ}$V          \\
                                &       & $\pm$1.5-5cm range precision                 \\
    \midrule
                                &       & 3-axis gyroscope                             \\
    XSens MTi-30 AHRS           & 200Hz & 3-axis accelerometer                         \\
                                &       & 3-axis magnetometer                          \\
    \midrule
    OptiTrack MoCap             & 120Hz & 6 DoF ground truth                           \\
    \midrule
    FARO Laser Scanner          & N/A   & accurate pointcloud scanning                 \\ 
    \bottomrule
  \end{tabular}
  \end{center}
\end{table}

The event stereo cameras have VGA resolution ($640\times480$) with a horizontal baseline of about 17cm. Since the MoCap system emits 850nm infrared strobes to locate the passive spherical markers, we have adopted the common practice of putting an infrared filter (PHTODE IR690, cutoff frequency of 400-690nm) in front of the lenses~\cite{klenk2021tum, calabrese2019dhp19}, thus blocking most of the flashing and reducing the background noise in the event stream. The reason for choosing a pair of event cameras with VGA ($640\times480$) rather than HD resolution ($1280\times720$) is due to the fact that we have observed a \emph{smearing effect} on top of the surface of active events. Similar problems are reported by Hu et~al.~\cite{hu2021v2e} and Alzugaray and Chli\cite{alzugaray2018asynchronous}, who make reference to motion blur in the event stream or timestamp delays owing to sudden and significant contrast changes on DAVIS event cameras. From an engineering perspective, the higher the number of events the more likely such artefacts occur. Besides camera dynamics, the resolution of the sensor is an obvious parameter deciding over the number of events.

The regular stereo cameras have global shutters and are positioned just under the event cameras with a vertical baseline of about 3.5cm. Owing to insufficient bandwidth, we have configured a $2\times2$ binning of pixels thus resulting in $1224\times1024$ monochrome frames. An additional \mbox{RGB-D} sensor is mounted in the middle and set up with unbinned narrow-field-of-view depth mode. The central mounting ensures that the overlap between the different cameras is maximized. Given the fact that the color camera on the sensor uses a rolling shutter mechanism, only the depth stream is recorded. We furthermore provide an easy-to-use function that uses the carefully-calibrated extrinsics to reproject the depth readings into other sensor frames.

A 128-channel panoramic LiDAR with a vertical opening angle of 90$^{\circ}$ and an azimuthal sampling density of 2048 beams per turn (the largest resolution commercially available) is placed on top of the setup. It is raised by four copper pillars to make sure the bottom beams are not blocked by the holder or any other sensors. Note that the \mbox{RGB-D} sensor and the LiDAR are unlikely to be used at the same time. The former is used in small-scale scenarios, while the latter operates best in large-scale scenarios.

All data sequences are recorded on a PC running Ubuntu 18.04 LTS on an Intel Core i9-10900F Processor. The computer is equipped with a GeForce GTX 1070 GPU, 32GB RAM, and a 1TB SSD. The details of each sensor are summarized in Table \ref{hw_spec_table}.

\subsection{Time Synchronization}

\begin{figure}[t]
  \centering
  \includegraphics[width=0.475\textwidth]{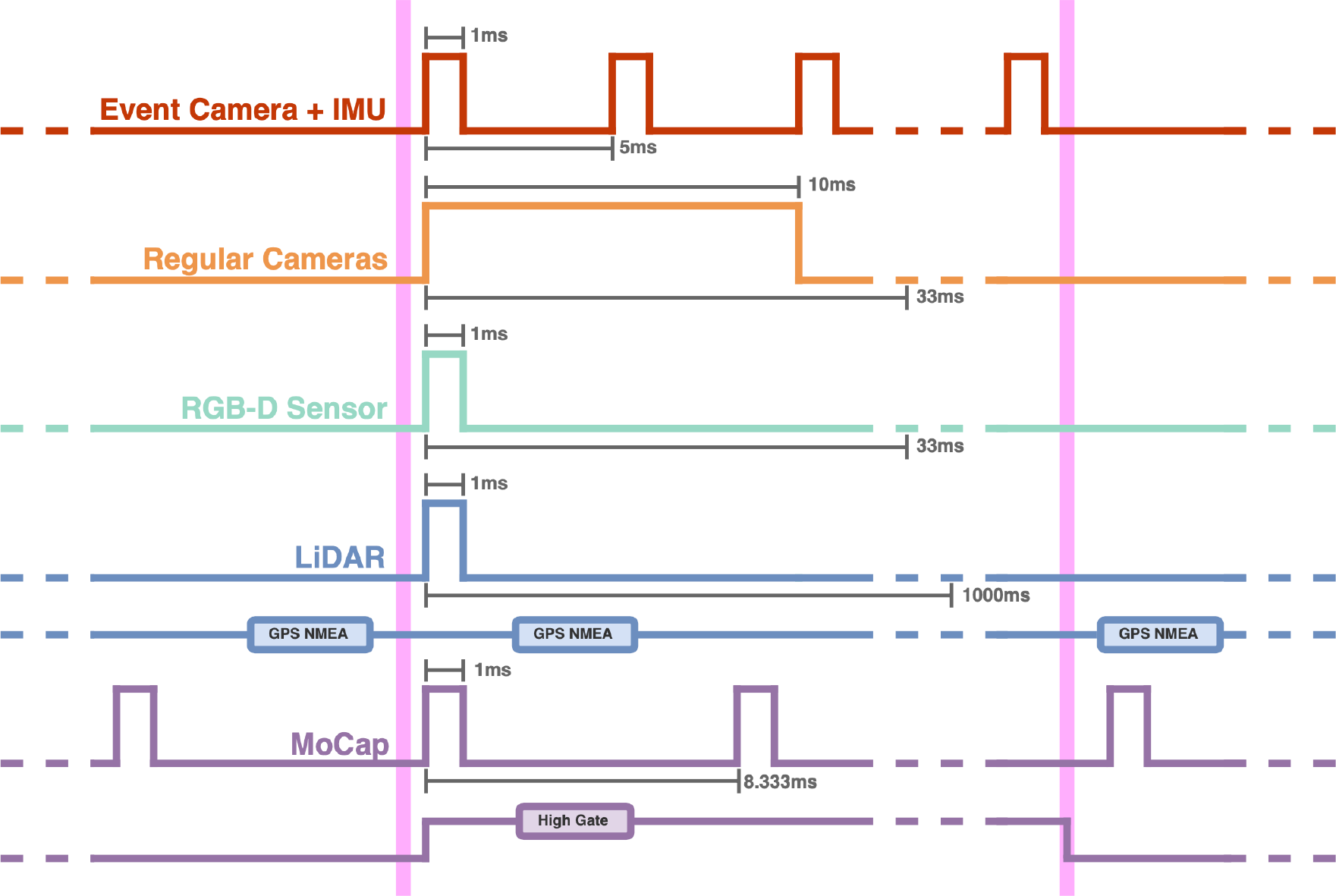}
  \caption{Illustration of the time synchronization implementation. An MCU produces different types of square wave signals. The different signals are marked with different colors, the period is indicated below the signal, and the pulse width above. The vertical lines on the left and right (purple) indicate the temporal location of the start and ending signals communicated to the MCU. Note that only one signal is indicated for each stereo pair.}
  \label{sync_fig}
\end{figure}

As illustrated in Fig.~\ref{sync_fig}, all sensors including the MoCap system are synchronized at the hardware level and triggered by a micro-controller unit (MCU). We use an STM32F407 and rely on the onboard external oscillator as a main clock to produce the trigger signals.

To suit the needs of the different types of sensors, the MCU outputs a variety of signals that adhere to the different synchronization interface specifications. After the MCU receives a start trigger from the user, it forwards a sequence of 200 pulses to the right event camera (master). This signal is used to calibrate the internal clock of the event camera in 5ms intervals, and synchronization within the pair is achieved via the internal firmware and a daisy chain connection. The same signal is also forwarded to the IMU. For the regular cameras, the MCU transmits a 30Hz signal with a 10ms pulse width to control the exposure intervals' beginning and end. The \mbox{RGB-D} sensor also receives a 30Hz signal but with a 1ms pulse width as required. The LiDAR's internal clock is synchronized with the MCU's through a mimicked analog GPS NMEA time signal, and individual scan capture is triggered by an additional 1Hz signal. The MoCap system receives a constant 120Hz signal from the MCU irrespectively of whether or not data recording is in progress. An independent high gate signal is used to mark the start and the end of an entire sequence. When the MCU receives an end trigger from the user, all aforementioned signals except for the constant MoCap signal and the GPS NMEA signal are terminated. The \mbox{RGB-D} sensor and the MoCap system both operate within the 800-900nm infrared band, which may cause cross-talk between the sensors. Fortunately, the infrared camera's exposure time is in the order of microseconds, and we insert a ghost camera in the MoCap interval scheduler to represent the \mbox{RGB-D} depth channel's actual exposure time interval.

We record all data streams into ROS bag files. Given that the hardware setup can achieve sub-microsecond synchronization accuracy, we use the order of each sensor's data packages to reconstruct the timeline, and cut off blank messages before and after the start and end trigger times.

\subsection{Calibration}

\begin{figure}[t]
  \centering
  \includegraphics[width=0.455\textwidth]{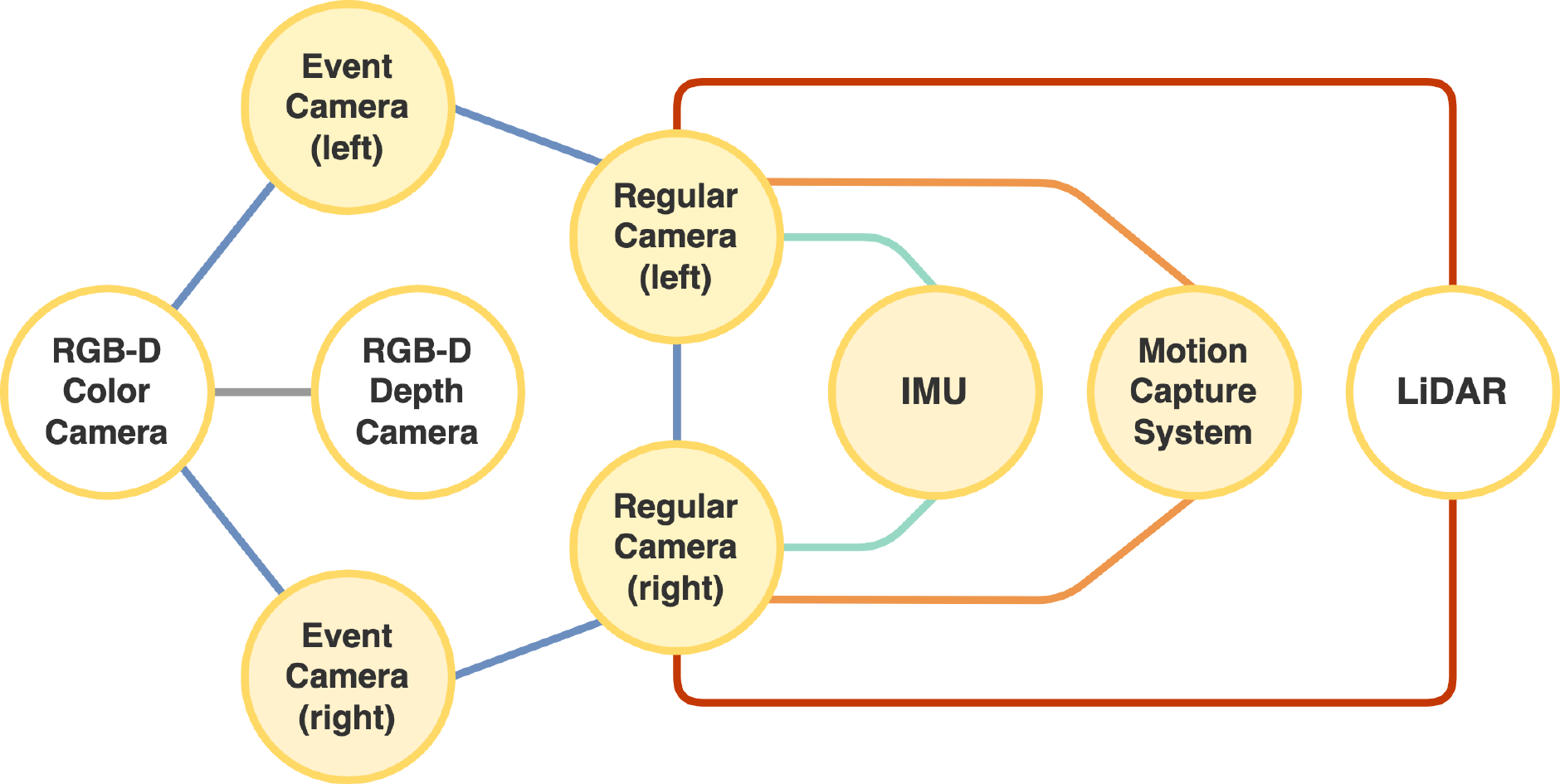}
  \caption{Illustration of required calibration variables. Nodes: sensor type (yellow/void background color indicates need/no need for intrinsic calibration). Blue edges: joint camera extrinsic calibration. Cyan edges: Camera-IMU extrinsic calibration. Orange edges: Camera-MoCap hand-eye calibration. Red edges: Camera-LiDAR extrinsic calibration. The variables are calibrated in the listed order.}
  \label{calib_fig}
\end{figure}

The whole sensor suite along with the MoCap reference frame requires intrinsic and extrinsic calibration. An illustration of all calibration variables and their order of calibration is shown in Fig.~\ref{calib_fig}. All data sequences used for calibration and the related software as well as the calibration results can be found on our website. For further reference, we also provide CAD measurements of all inter-sensor extrinsic parameters.

\subsubsection{Intrinsics}

All camera intrinsics, including the focal lengths ($f_x$, $f_y$), the optical center ($c_x$, $c_y$), and the distortion parameters ($k_1$, $k_2$, $p_1$, $p_2$), are calibrated using the official ROS camera calibration toolbox\footnote{\url{http://wiki.ros.org/camera_calibration}} and by gently moving in front of a $9\times6$ checkerboard visualized on a computer screen. The choice of a virtual checkerboard enables display in either \emph{static} or \emph{blinking} mode. The latter is particularly useful for event camera calibration, as it produces accumulated event images with a sharp appearance of the checkerboard. Images that contain too much blur have been manually removed. Note that we take the factory calibration result for the \mbox{RGB-D} sensor (intrinsics of both color and depth camera as well as extrinsics between them).
The IMU intrinsics (i.e. the statistical properties of the accelerometer and gyroscope signals, including bias random walk and noise densities) are calibrated using the Allan Variance ROS toolbox\footnote{\url{https://github.com/ori-drs/allan_variance_ros}} with a 5-hour-long IMU sequence by putting the sensor flat on the ground with no perturbation. The MoCap system is pre-calibrated before data recording.

\subsubsection{Joint camera extrinsic calibration}

In order to determine the extrinsics of the multi-camera system, we point the sensor setup towards the screen and record both static and blinking checkerboard patterns with known size. For each observation, the relative position between screen and sensors is kept still by putting the sensor suite steadily on a tripod. The board is maintained within the field of view of all cameras. Note that here we use the color camera on the \mbox{RGB-D} sensor to jointly calibrate its extrinsics. The extrinsics of the depth camera are obtained by the known internal parameters of the depth camera, and the rolling shutter effect is safely ignored as no motion between cameras and pattern is involved. The extrinsics are calculated by detecting corner points on the checkerboard pattern and applying PnP to the resulting 2D-3D correspondences. The result is refined by Ceres Solver\footnote{\url{http://ceres-solver.org/}}-based reprojection error minimization. We finally validate the estimated extrinsic parameters by analyzing the quality of depth map reprojections and by comparing the result against the measurements from the CAD model.

\subsubsection{Camera-IMU extrinsic calibration}

Extrinsic transformation parameters between the IMU and the regular stereo camera are identified using the Kalibr toolbox~\cite{furgale2013unified, furgale2012continuous}.
The visual-inertial system is directed towards a static $6\times6$ April-grid board, and the board is constantly maintained within the field of view of both regular cameras. All six axes of the IMU are properly excited, and the calibration is conducted under good illumination conditions to further reduce the unwanted side-effects of motion blur. Given prior intrinsics of the regular stereo camera and the IMU and extrinsics between the regular cameras, we limit the calculation to the extrinsics between the IMU and the regular stereo camera, only.

\subsubsection{Camera-MoCap hand-eye calibration}

The MoCap system outputs position measurements of the geometric centers of all markers expressed in a MoCap-specific reference frame. In order to compare recovered trajectories against ground truth, we therefore need to identify a euclidean transformation between the MoCap frame of reference and any other sensor frame. We follow the hand-eye calibration paradigm presented in~\cite{daniilidis1999hand}. A static $7\times6$ checkerboard is maintained within the field of view of both gently-moving cameras, and MoCap pose measurements are simultaneously recorded. Relative poses from both the MoCap system and the cameras are then used to solve the hand-eye calibration problem using the official OpenCV calibrateHandEye API.\footnote{\url{https://docs.opencv.org/4.2.0/d9/d0c/group__calib3d.html}}

\subsubsection{Camera-LiDAR extrinsic calibration}

Our extrinsic calibration between the LiDAR and the cameras bypasses via a high-quality colored point cloud captured by a FARO laser scanner. The point cloud is captured in an unfurnished room with simple geometric structure and in which we only place a checkerboard. In order to perform the extrinsic calibration, we then record LiDAR scans and corresponding camera images by moving the sensor setup in this room. The cameras are constantly directed at the checkerboard. Next, we estimate the transformation between the FARO and the LiDAR coordinate frames by point cloud registration. Owing to the fact that the FARO scan is very dense and colored, we can furthermore hand-pick 3D points corresponding to checkerboard corners in the real world. By furthermore detecting those points in the camera images, we can again run the PnP method to obtain FARO to camera transformations. To conclude, the extrinsic parameters between LiDAR and cameras are retrieved by concatenating the above two transformations.


\section{Dataset Overview}

\begin{table}[t]
  \caption{Bias settings for the event cameras}
  \label{bias_table}
  \begin{center}
  \begin{tabular}{cccc}
    \toprule
    bias param. & well-lit & low-light & sunlight \\
    \midrule
    diff        & 299      & 299       & 299      \\
    diff\_off   & 221      & 200       & 221      \\
    diff\_on    & 384      & 420       & 384      \\
    fo          & 1598     & 1530      & 1550     \\
    hpf         & 1437     & 1530      & 1448     \\
    pr          & 1250     & 1250      & 1250     \\
    refr        & 1500     & 1500      & 1500     \\
    \bottomrule
  \end{tabular}
  \end{center}
\end{table}

All sensors are mounted on a versatile 3D-printed holder which is resistant to deformations and can be mounted on various platforms: (1) a simple handle for hand-held sequences, (2) a wheeled tripod, and (3) a helmet. Overall, the datasets can be divided into two categories: \emph{small-scale} and \emph{large-scale}. The small-scale data sequences are recorded inside the MoCap arena, which has a volume of $5m\times5m\times3m$. On small-scale sequences, all sensors are recorded except for the panoramic LiDAR. Large-scale data sequences are recorded by exerting longer trajectories through larger-scale indoor architectures. All sensors except for the \mbox{RGB-D} sensor are recorded. We again use the FARO laser scanner to generate high-quality point clouds of the environment, and align LiDAR scans to this point cloud in order to generate ground truth information. Each sequence is provided in the form of a set of individual, single-topic ROS bags, and the ground truth signal is provided as a separate file. All bags can be downloaded from our website, and the user may select a subset for each sequence depending on the sensor setup. We provide an additional, easy-to-use dataset toolbox to execute functions such as calibration, event visualization, bag merging, data validation, and depth reprojection. Sensor measurements and scene setups for a selection of four sequences is presented in Fig.~\ref{data_seq_fig}.

In order to control the number of events and ensure a clean observation of the texture and the structure of the environment, and the sensitivity of the event camera (i.e. the biases\footnote{\url{https://docs.prophesee.ai/stable/hw/manuals/biases.html}}) is carefully tuned. Table~\ref{bias_table} summarizes our settings. For small-scale data sequences captured in a constantly well-lit environment (about 300lux, high SNR), we chose a conservative set of bias parameters (i.e. \emph{well-lit}) and the smallest aperture in order to maximize the suppression of background noise and reduce the overall number of generated events. As a result, the captured event stream is generally more distinctive. In extremely low light situations (around 1lux, low SNR), the aperture is adjusted to the largest setting, and a different set of bias parameters is chosen to increase the event camera's responsiveness to brightness changes (i.e. \emph{low-light}). For large-scale data sequences, we adopt one of two sets of bias parameters (\emph{well-lit} or \emph{sunlight}), and the choice highly depends on how much natural light is present in the scene.

\subsection{Small-scale sequences}

\begin{table}[t]
  \caption{Overview of small-scale data sequences}
  \label{small_scale_data_seq_table}
  \begin{center}
  \begin{tabular}{ccccc}
    \toprule
    Seq. Name              & Time[s] & MER[$10^6$events/s] & Description      \\ 
    \midrule
    \emph{board-slow}      & 35      & 0.86                & Planar Motion    \\ 
    \emph{corner-slow}     & 41      & 0.67                & Spherical Motion \\ 
    \emph{robot-normal}    & 40      & 0.54                & 6 DoF            \\ 
    \emph{robot-fast}      & 31      & 3.02                & 6 DoF            \\ 
    \emph{desk-normal}     & 91      & 1.07                & 6 DoF            \\ 
    \emph{desk-fast}       & 46      & 7.14                & 6 DoF            \\ 
    \emph{sofa-normal}     & 91      & 2.02                & 6 DoF            \\ 
    \emph{sofa-fast}       & 40      & 4.18                & 6 DoF            \\ 
    \emph{mountain-normal} & 61      & 4.62                & 6 DoF            \\ 
    \emph{mountain-fast}   & 43      & 13.09               & 6 DoF            \\ 
    \emph{hdr-normal}      & 60      & 2.28                & 6 DoF, low light \\ 
    \emph{hdr-fast}        & 41      & 10.37               & 6 DoF, low light \\ 
    \bottomrule
  \end{tabular}
  \end{center}
\end{table}

All small-scale data sequences are summarized in Table~\ref{small_scale_data_seq_table}. The name of each sequence indicates the observed scene, and the postfix indicates the motion speed. \emph{MER} represents the mean event rate of the left event camera, omitting the first and last 5s of recording. All sequences are recorded within the boundaries of the MoCap arena, and contain multiple closed loops on each trajectory. The length of each sequence is indicated in seconds. The first two sequences are well-suited for debugging purposes. \emph{board-slow} contains three DoF planar motion, pointing at a flat, texture-rich paperboard and a 3D geometric object. \emph{corner-slow} contains a three DoF spherical motion by rotating the ballhead of the tripod. The setup faces a closet with checkerboard texture and further box-shaped objects positioned in a nearby corner. The setup is purposely built to contain large segments with different, continuous depths. All remaining data sequences contain full six DoF motion. The \emph{robot} sequences show a simple scene with a humanoid robot placed on a white table. The \emph{desk} sequences mimic a messy working environment with multiple books, on-and-off screens, keyboards, headphones, and other accessories and decorations placed on an L-shaped table with a chair placed in front of it. The \emph{sofa} sequences show a typical living room environment with a large sofa surrounded by multiple chairs and a coffee table. Some extra items are distributed on the furniture to enrich the texture. The \emph{mountain} sets are composed of several calibration targets, thus resulting in feature-rich but self-similar texture. Each board is placed in a random position causing occlusions in the data. To conclude, the \emph{hdr} sequences are captured in low light conditions. The setup is similar to the \emph{mountain} sequences, except that the camera directly faces into a light source.

\subsection{Large-scale sequences}

\begin{table}[t]
  \caption{Overview of large-scale data sequences}
  \label{large_scale_seq_table} 
  \begin{center}
    \begin{tabular}{cccc}
      \toprule
      Seq. Name              & Length[m] & MER[$10^6$events/s] & Description \\ 
      \midrule
      \emph{corridors-dolly} & 80        & 0.58                & One Loop    \\ 
      \emph{corridors-walk}  & 80        & 0.85                & One Loop    \\ 
      \emph{units-dolly}     & 244       & 1.01                & Loops       \\ 
      \emph{units-scooter}   & 241       & 1.83                & Loops       \\ 
      \emph{school-dolly}    & 119       & 1.61                & One-way     \\ 
      \emph{school-scooter}  & 111       & 2.80                & One-way     \\ 
      \bottomrule
    \end{tabular}
  \end{center}
\end{table}

The large-scale data sequences are recorded in various indoor environments with long trajectories, as summarized in Table~\ref{large_scale_seq_table}. Owing to different platforms, the sequences capture three different types of motion. The platforms are indicated by the postfix of each name: (1) a dolly, thus resulting in three DoF planar motion; (2) a helmet, thus resulting in jerky, full six DoF walking trajectories; and (3) a helmet worn by a scootering person, thus resulting in smooth, full six DoF high-speed motion. The \emph{corridor} sequences are recorded in the basement of ShanghaiTech University's teaching center, and consist of four corridors connected in a loopy Q-shape. The environment contains scarce texture and approximately constant, well-lit illumination without the influx of natural sunlight. The \emph{units} sequences are recorded during nighttime, and present various sub-scenes reaching from an empty school to a cluttered room and a long, straight hallway. The sensors are deliberately moved in and out of units to form multiple closed loops. The \emph{school} sequences are captured along non-loopy trajectories on a single floor of a school building. \emph{school} is recorded in well-furnished environments with lots of texture. \emph{corridor} and \emph{units} use the \emph{well-lit} set of bias parameters, while \emph{school} is recorded during daytime with incoming sunlight, and thus again present HDR conditions. We therefore use the \emph{sunlight} bias on those sequences.


\section{EVALUATION}

\subsubsection{Ground Truth}

For small-scale data sequences, ground truth is given by applying the hand-eye calibration result to the MoCap readings. The latter have sub-millimeter accuracy, a frequency of 120Hz, and six DoF. For large-scale data sequences, we follow the approach proposed by Ramezani et~al.~\cite{ramezani2020newer} and use the Iterative Closest Point (ICP) method to match the motion-compensated LiDAR scans with a pre-scanned dense point cloud of the environment captured by the survey-grade FARO laser scanner, thus resulting in a six DoF ground truth trajectory for the LiDAR frame. We use CloudCompare\footnote{\url{https://www.danielgm.net/cc/}}, a robust and efficient open-source software to register the LiDAR scans with the high-resolution prior global map. The procedure merely requires an initial guess for the first scan in each sequence, which we set manually. The initial guess of each subsequent scan is then simply set to the optimized pose of the previous scan, thus leading to  the complete ground truth trajectory.

\subsubsection{Evaluation metrics}

To validate and evaluate the performance of different algorithms, the provided scripts calculate Relative Pose Errors (RPE) and Absolute Trajectory Errors (ATE) as defined in the original work of Sturm et~al.~\cite{sturm2012benchmark}. These measures express local tracking accuracy and global drift and consistency, respectively. We use the opensource software EVO\footnote{\url{https://github.com/MichaelGrupp/evo}} to perform the calculation and comparison. Note that before we compare the ground truth trajectory against the trajectory generated by an algorithm, we express the ground truth results into the respective sensor's frame using our setup's extrinsic parameters. Given that SLAM algorithms express poses with respect to an arbitrary initial starting frame, we furthermore use the first N poses of each trajectory to align the results with ground truth.

\subsubsection{Validation on state-of-the-art algorithms}

\begin{table*}[t]
  \caption{State-Of-The-Art algorithms' Performance} 
  \label{evaluation_table}
  \begin{center}
    \begin{tabular}{ccccccccccccc}
      \toprule
                       & \multicolumn{3}{c}{\underline{\smash{\emph{\ \ \ \ \ \ \ \ \ \ desk-normal\ \ \ \ \ \ \ \ \ \ \ \ }}}}
                       & \multicolumn{3}{c}{\underline{\smash{\emph{\ \ \ \ \ \ \ \ \ robot-normal\ \ \ \ \ \ \ \ \ \ \ \ }}}}
                       & \multicolumn{3}{c}{\underline{\smash{\emph{\ \ \ \ \ \ \ \ corridors-walk\ \ \ \ \ \ \ \ }}}}
                       & \multicolumn{3}{c}{\underline{\smash{\emph{\ \ \ \ \ \ \ \ \ \ units-dolly\ \ \ \ \ \ \ \ \ \ \ }}}} \\
                       & $\mathbf{R}_{\text{rpe}}$ & $\mathbf{t}_{\text{rpe}}$[cm] & $\mathbf{t}_{\text{ate}}$[cm]
                       & $\mathbf{R}_{\text{rpe}}$ & $\mathbf{t}_{\text{rpe}}$[cm] & $\mathbf{t}_{\text{ate}}$[cm]
                       & $\mathbf{R}_{\text{rpe}}$ & $\mathbf{t}_{\text{rpe}}$[m]  & $\mathbf{t}_{\text{ate}}$[m]
                       & $\mathbf{R}_{\text{rpe}}$ & $\mathbf{t}_{\text{rpe}}$[m]  & $\mathbf{t}_{\text{ate}}$[m] \\
      \midrule
      ORB-SLAM3-Stereo & 0.0032 & 0.23 &  5.38 & 0.0089 & 0.41 & 4.32 & 0.0272 & 0.16 & 2.43 & 0.0174 & 0.27 & 2.86 \\
      ORB-SLAM3-RGBD   & 0.0028 & 0.23 &  6.36 & 0.0055 & 0.21 & 7.29 & -      & -    & -    & -      & -    & -    \\
      VINS-mono        & 0.0035 & 0.60 & 25.03 & 0.0064 & 0.37 & 3.52 & 0.0380 & 0.10 & 1.18 & 0.0231 & 0.17 & 8.93 \\
      VINS-fusion      & 0.0034 & 0.42 & 10.48 & 0.0074 & 0.35 & 3.86 & 0.0318 & 0.08 & 1.24 & 0.0324 & 0.08 & 2.37 \\
      LIO-SAM          & -      & -    & -     & -      & -    & -    & 0.0104 & 0.04 & 0.26 & 0.0153 & 0.06 & 0.40 \\ 
      \bottomrule
    \end{tabular}
  \end{center}
\end{table*}

\begin{figure*}[t]
  \centering
  \includegraphics[width=\textwidth]{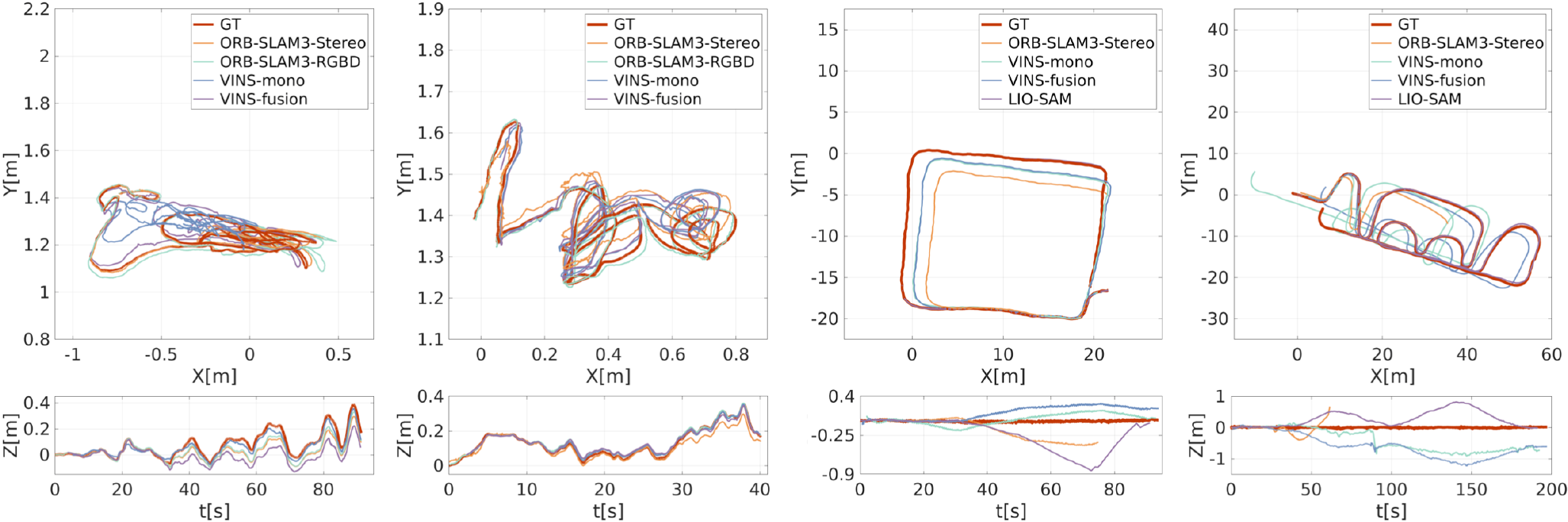}
  \caption{From left to right: \emph{desk-normal}, \emph{robot-normal}, \emph{corridors-walk}, \emph{units-dolly}. From top to bottom: bird's-eye view of trajectories, details along Z-axis.}
  \label{traj_fig}
\end{figure*}

We validated our data sequences by applying ORB-SLAM3~\cite{campos2021orb} with regular stereo setting, ORB-SLAM3 with \mbox{RGB-D} setting, VINS-mono~\cite{qin2018vins} with monocular-inertial setting, VINS-fusion with stereo-inertial setting, and LIO-SAM~\cite{liosam2020shan} with LiDAR-inertial setting. As shown in Table \ref{evaluation_table} and Fig.~\ref{traj_fig}, the results obtained by the above frameworks are in line with their expected performance, which validates our dataset preparations and suggests high practical usefulness. The current level of existing event-based SLAM methods, as documented on our website, confirms the relevance of the presently-proposed development and benchmark kit, which further demonstrates the open challenges and needs for novel event-based algorithms.


\section{CONCLUSIONS}

In conclusion, we are proposing novel benchmark datasets for research on multi-sensor SLAM extending the commonly used sensor set by dynamic vision sensors. The datasets are fully synchronized and accurately calibrated, and we expect them to be of high value to the community. Each dataset has been recorded multiple times with slightly different trajectories. Our current work consists of implementing an open SLAM benchmark webpage that accepts public algorithm submissions. It will use the unpublished, secret sequences to perform a fully automatic, fair evaluation and ranking of any submitted algorithm.



\end{document}